\documentclass[10pt]{article}
\usepackage{iclr2025_conference,times}
\usepackage{amsmath,amssymb}
\usepackage{graphicx}
\usepackage{booktabs}
\usepackage{natbib}
\usepackage{hyperref}
\usepackage{cleveref}

\iclrfinalcopy

\title{A Scaling Law for Token Efficiency in LLM Fine-Tuning Under Fixed Compute Budgets}

\author{Ryan Lagasse, Aidan Kierans, Avijit Ghosh \& Shiri Dori-Hacohen\\
University of Connecticut\\
Storrs, CT 06226, USA \\
\texttt{ryan.lagasse@uconn.edu} \\
}

\begin{document}

\maketitle

\begin{abstract}
We introduce a scaling law for fine-tuning large language models (LLMs) under fixed compute budgets that explicitly accounts for data composition. Conventional approaches measure training data solely by total tokens, yet the number of examples and their average token length---what we term \emph{dataset volume}---play a decisive role in model performance. Our formulation,
\[
\text{Accuracy} = A\, V^{\beta} M^{\gamma} + E,
\]
where \textit{Volume} \(V=N\cdot L\) (number of examples $\times$ average token length) and \(M\) is the model size, is tuned following established procedures \citep{pareja2024unveilingsecretrecipeguide}. Experiments on the BRICC dataset \cite{salavati2024reducing} and subsets of the MMLU dataset \cite{hendrycks2021measuringmassivemultitasklanguage}, evaluated under multiple subsampling strategies, reveal that data composition significantly affects token efficiency. These results motivate refined scaling laws for practical LLM fine-tuning in resource-constrained settings. Code will be made available upon publication.

\end{abstract}

\section{Introduction}
Scaling laws have emerged as a powerful tool for predicting performance in large-scale neural networks, as demonstrated in recent work \citep{hernandez2021scalinglawstransfer, hoffmann2022trainingcomputeoptimallargelanguage} and in fine-tuning \cite{zhang2024scalingmeetsllmfinetuning}. Yet in the context of fine-tuning LLMs for domain-specific applications, these laws often reduce training data to a single metric (total tokens) while neglecting the inherent compositional differences in the data. In real-world scenarios, practitioners face not only limited data but also data whose structure---in terms of the number of examples versus their individual lengths---varies considerably. Two datasets with identical total tokens may yield drastically different performance if one contains many short examples and the other a few long ones. Our work thereby extends existing scaling laws to more accurately capture practical fine-tuning dynamics under fixed compute constraints.

\section{Methodology}
Our approach redefines the effective data size by decomposing the total token count into the number of examples \(N\) and their average token length \(L\). Although \(V=N\cdot L\) is mathematically equivalent to total tokens, it explicitly emphasizes data composition. We hypothesize that fine-tuning accuracy scales as
\[
\text{Accuracy} = A\, V^{\beta} M^{\gamma} + E,
\]
where \(A\), \(\beta\), \(\gamma\), and \(E\) are tuned parameters (see \citealp{zhang2024scalingmeetsllmfinetuning} for a similar approach). In our experiments we consider three distinct subsampling strategies---\texttt{few\_long}, \texttt{many\_short}, and \texttt{balanced}---to isolate the impact of data composition on performance. We evaluate on the BRICC dataset using models of sizes 135M, 360M, 500M, and 1B, under a fixed compute budget from the SMOLLM, QWEN, and Falcon families that perform best on the Open LLM Leaderboard (\cite{open-llm-leaderboard}) for their respective model sizes. For brevity, the details of the parameter tuning, models, and fitting procedure follow established methods and are omitted here and included in the appendix.

\section{Experiments}
\begin{figure}[ht]
\centering
\begin{minipage}{0.48\textwidth}
    \centering
    \includegraphics[width=\textwidth]{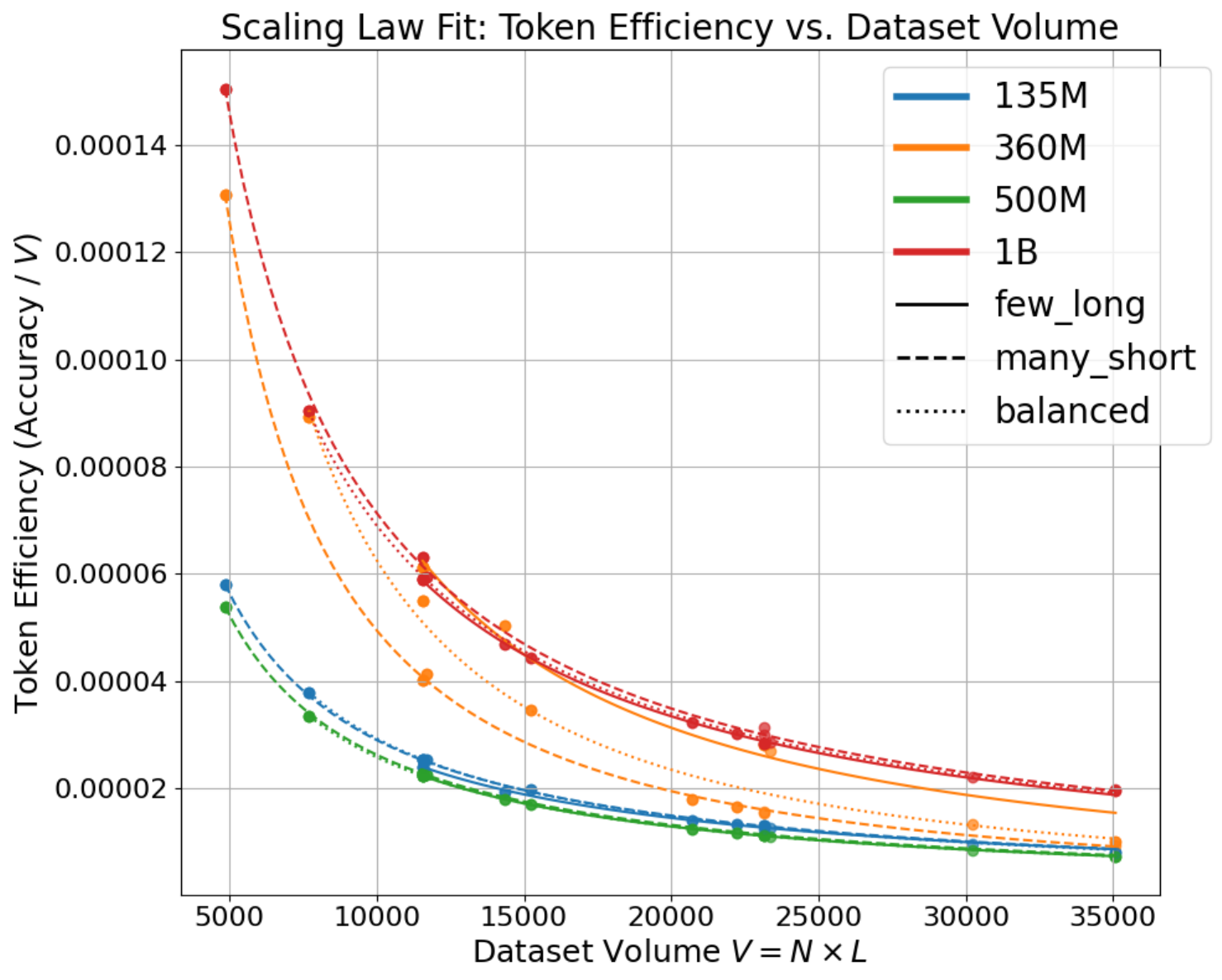}
    \caption{Normalized token efficiency \(\eta_{\mathrm{norm}}=(\text{Accuracy}-E)/(V\,M^{\gamma})\) as a function of model size. The trend indicates that larger models exhibit superior token efficiency when data composition is properly accounted for.}
    \label{fig:boxplot2}
\end{minipage}
\hfill  
\begin{minipage}{0.48\textwidth}
    \centering
    \includegraphics[width=\textwidth]{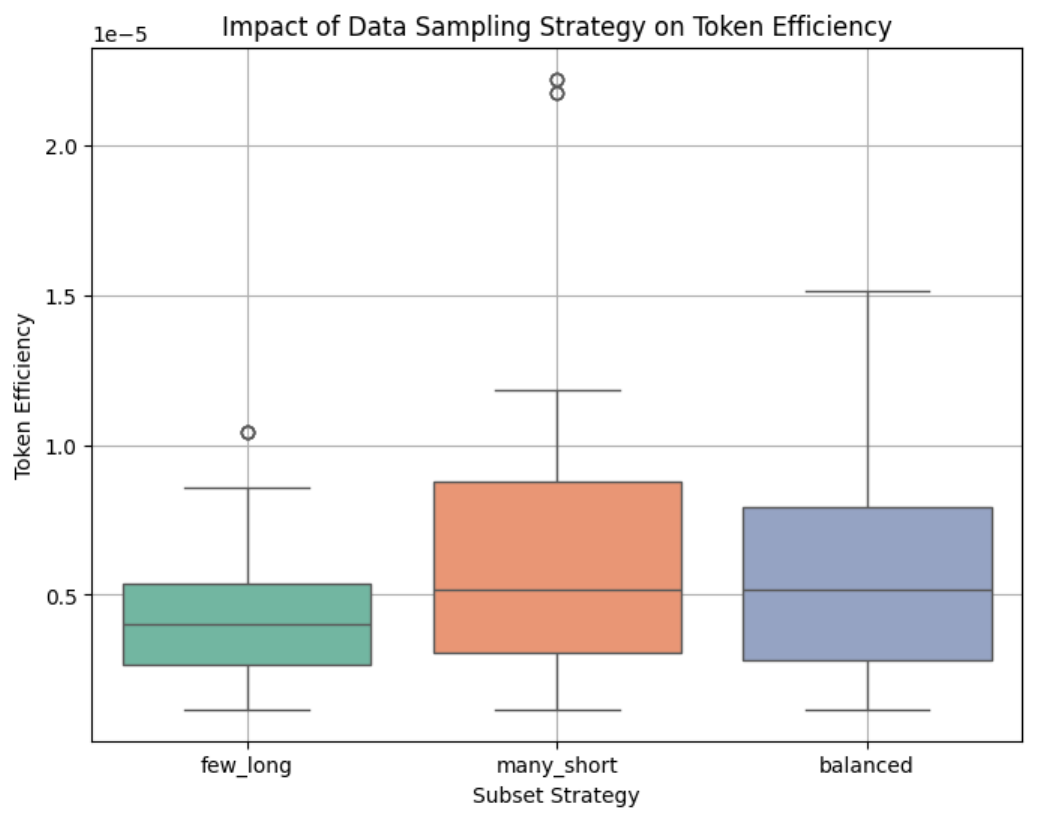}
    \caption{Box plots of accuracy for each subsampling strategy, demonstrating the influence of data composition on performance. The distinct medians indicate that subsampling strategy is a critical factor in fine-tuning outcomes.}
    \label{fig:efficiency2}
\end{minipage}
\end{figure}

We evaluate our scaling law on the BRICC dataset and subsets of the MMLU using models of sizes 135M, 360M, 500M, and 1B. Experiments were performed under a fixed compute budget with three subsampling strategies. Table~\ref{tab:results} summarizes representative performance metrics and the corresponding dataset volume for each strategy. Our analysis reveals that, for a given model size, performance variations are closely correlated with differences in \(V\). Figure~\ref{fig:boxplot2} illustrates the distribution of accuracy across subsampling strategies, underscoring the influence of data composition. Figure~\ref{fig:efficiency2} presents the normalized token efficiency, defined as
\[
\eta_{\mathrm{norm}}=\frac{\text{Accuracy}-E}{V\,M^{\gamma}},
\]
which demonstrates that, when properly normalized, larger models leverage additional tokens more effectively.

\begin{table}[ht]
\centering
\caption{Representative performance on BRIMI for various subsampling strategies. The dataset volume \(V\) is computed as the product of the number of examples and the average token length.}
\label{tab:results}
\begin{tabular}{lcccc}
\toprule
Strategy & Mean \(N\) & Mean \(L\) & \(V\) & Mean Accuracy\\
\midrule
few\_long   & 171  & 67.6 & 11558 & 0.278\\
many\_short & 455  & 25.4 & 11570 & 0.294\\
balanced    & 347  & 33.3 & 11545 & 0.294\\
\bottomrule
\end{tabular}
\end{table}

\section{Discussion and Conclusion}
Our results clearly demonstrate that incorporating data composition via the dataset volume \(V=N\cdot L\) yields a scaling law that predicts fine-tuning performance under fixed compute budgets. The observed variations across subsampling strategies confirm that the nature of the training data—not merely its total token count—critically influences token efficiency. Although the tuning procedure for the scaling parameters follows that of prior work, our formulation provides a much better perspective into important problems with small data and GPU limits, which are realistic limiters for many researches and are essential for optimizing LLM fine-tuning in practical, resource-constrained environments.

In summary, by capturing the interplay between dataset composition and model size, our scaling law framework offers actionable insights for practitioners and lays the groundwork for future extensions, including those to quantized and parameter-efficient training regimes.

\bibliographystyle{iclr2025_conference}
\bibliography{iclr2025_conference}

\appendix
\section*{Appendix}

\subsection*{BRICC Dataset}
The BRICC dataset \cite{salavati2024reducing} is a proprietary, domain-specific benchmark designed to assess fine-tuning performance under real-world, low-resource conditions. It consists of 1,530 annotated text segments extracted from [domain-specific sources, e.g., financial documents or customer inquiries]. The dataset was curated specifically for gender bias detection, featuring a 1:4 split that is skewed toward the bias class. The choice of BRICC is motivated by its relevance to specialized applications and by the fact that none of the models we use have been pre-trained on this data, ensuring a rigorous test of transfer performance.

\subsection*{MMLU Results}
To further validate our proposed scaling law and the importance of data composition, we conducted additional experiments on subsets of the MMLU dataset \citep{hendrycks2021measuringmassivemultitasklanguage}. These experiments followed the same setup as for the BRICC dataset, employing the three subsampling strategies (\texttt{few\_long}, \texttt{many\_short}, and \texttt{balanced}) and evaluating models with sizes of 135M, 360M, 500M, and 1B under a fixed compute budget.

\begin{figure}[ht]
\centering
\begin{minipage}{0.48\textwidth}
    \centering
    \includegraphics[width=\textwidth]{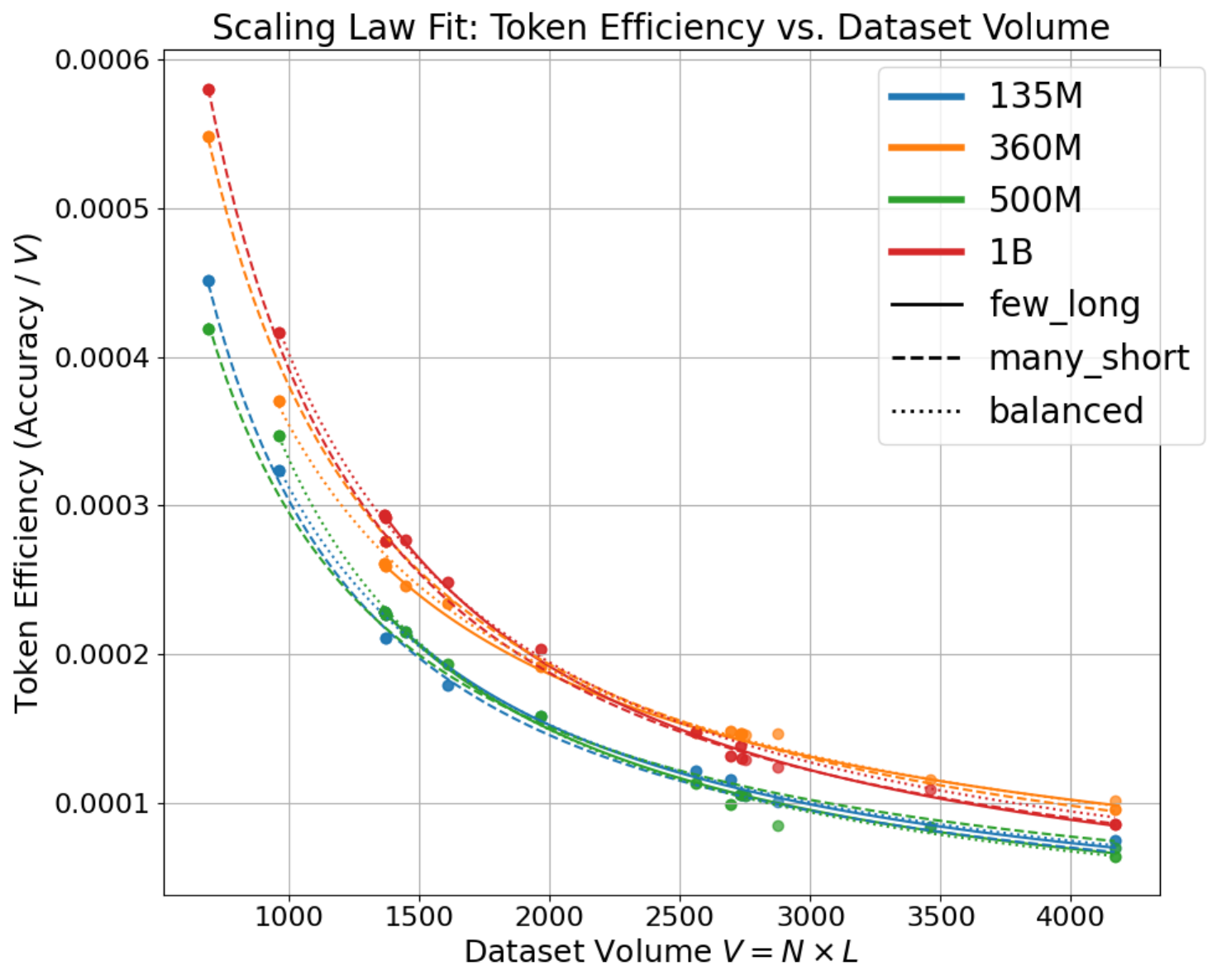}
    \caption{Normalized token efficiency as a function of model size now shown for the MMLU human aging dataset.}
    \label{fig:boxplot3}
\end{minipage}
\hfill  
\begin{minipage}{0.48\textwidth}
    \centering
    \includegraphics[width=\textwidth]{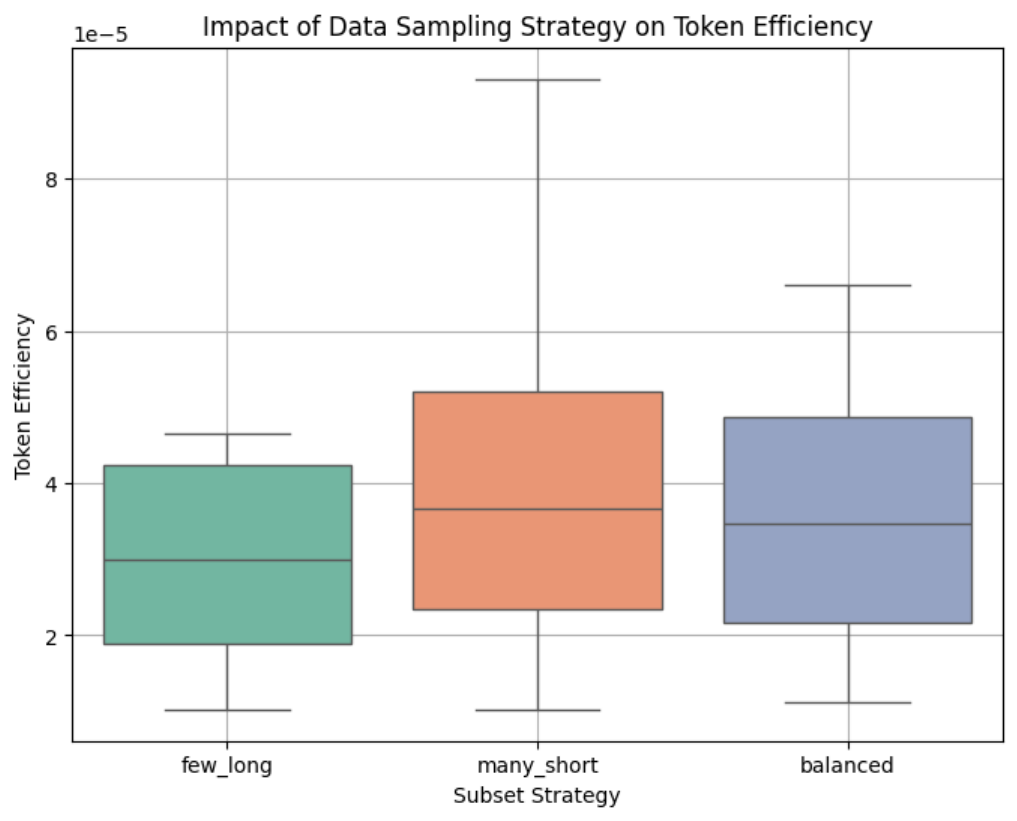}
    \caption{Box plots of accuracy for each subsampling strategy showing the same trend as BRICC. We tested many of the other MMLU tracks and found this is consistent.}
    \label{fig:efficiency3}
\end{minipage}
\end{figure}

Figure~\ref{fig:boxplot3} shows the normalized token efficiency, defined as
\[
\eta_{\mathrm{norm}} = \frac{\text{Accuracy}-E}{V\,M^{\gamma}},
\]
plotted as a function of model size. The trend evident in this figure indicates that, when data composition is properly accounted for through the dataset volume \(V = N \cdot L\), larger models exhibit superior token efficiency. This is consistent with our hypothesis that model performance is not merely a function of total tokens, but critically depends on how those tokens are composed into examples.

\begin{table}[ht]
\centering
\caption{Representative performance on MMLU for various subsampling strategies. The dataset volume \(V\) is computed as the product of the number of examples and the average token length.}
\label{tab:results}
\begin{tabular}{lcccc}
\toprule
Strategy & Mean \(N\) & Mean \(L\) & \(V\) & Mean Accuracy\\
\midrule
few\_long   & 77.91  & 30.17 & 11558 & 0.33\\
many\_short & 93.58  & 18.34 & 11570 & 0.34\\
balanced    & 88.50  & 21.94 & 11545 & 0.34\\
\bottomrule
\end{tabular}
\end{table}

Figure~\ref{fig:boxplot3} presents box plots of accuracy for each subsampling strategy. The distinct medians and variability among the strategies underscore that the structure of the data—whether it is dominated by many short examples, few long examples, or a balanced mix—significantly influences fine-tuning outcomes. In particular, the differences in the distributions confirm that two datasets with similar total token counts but differing compositions can yield markedly different performance, further reinforcing the need for our dataset volume formulation.

Together, these MMLU results complement our findings on the BRICC dataset and provide robust evidence that incorporating data composition into scaling laws is essential for accurately predicting LLM fine-tuning performance under resource constraints.

\subsection*{Subsampling Strategies}
Our experiments employ three subsampling strategies to probe the effect of data composition:
\begin{itemize}
\item \textbf{few\_long}: A strategy that selects a relatively small number of examples with long token lengths to maximize for token length in selection while trying to reach the maximum token threshold.
\item \textbf{many\_short}: A strategy that selects a large number of examples with short token lengths by maximizing the number of examples selected while staying under the maximum token limit.
\item \textbf{balanced}: A strategy that maintains a balance between the number of examples and their token lengths by picking examples close to the median token length.
\end{itemize}
Importantly all of these selected from the same dataset and as we increased the maximum token limit for models they contained all the same data as the previous runs. Detailed configurations and experimental settings are provided in our supplementary materials and in the code.

\subsection*{Additional Experimental Details}
The scaling law parameters \(A\), \(\beta\), \(\gamma\), and \(E\) are tuned following the procedure outlined in \cite{zhang2024scalingmeetsllmfinetuning}. Although the internal parameter values are not reported here, the tuning procedure is fully consistent with established practice.

\subsection*{Subsampling Strategies and Data Selection}
To probe the impact of data composition, we implement three distinct subsampling strategies. The \texttt{few\_long} strategy selects a relatively small number of examples that have long token lengths; the \texttt{many\_short} strategy selects a large number of examples with short token lengths; and the \texttt{balanced} strategy maintains an equilibrium between example count and token length. The goal of this was to cover the range of realistic data compositions encountered in practice. For each strategy, our data preprocessing pipeline (implemented in Python using Pandas and NumPy) computes \(N\), \(L\), and hence \(V\) for each experimental condition.

\subsection*{Experimental Design and Baseline Comparisons}
We evaluate our scaling law on the BRICC dataset and selected subsets of the MMLU dataset - specifically human aging, professional law, and anatomy, using four model sizes (135M, 360M, 500M, and 1B) chosen based on their performance on the HuggingFace LLM Leaderboard \citep{open-llm-leaderboard}. Our experimental design includes:
\begin{itemize}
    \item Ablation studies where we vary \(N\) and \(L\) independently while keeping \(V\) constant, confirming that both components contribute significantly to fine-tuning performance.
    \item Baseline comparisons with conventional scaling laws that use total token count, demonstrating that our formulation (using \(V\)) achieves lower prediction errors and better statistical significance.
\end{itemize}

\subsection*{Parameter Tuning Procedure}
Our scaling law is given by
\[
\text{Accuracy} = A\, V^{\beta} M^{\gamma} + E,
\]
where \(M\) denotes model size and the parameters \(A\), \(\beta\), \(\gamma\), and \(E\) are tuned using procedures analogous to those in \cite{zhang2024scalingmeetsllmfinetuning}. In our implementation, we performed a grid search over plausible values of the irreducible performance offset \(E\) (ranging from 0.20 to 0.30) to ensure that \(\text{Accuracy} - E\) remains positive for all data points. Once \(E\) is selected for each subsampling strategy, the model is linearized via the transformation
\[
\ln(\text{Accuracy} - E) = \ln A + \beta\, \ln(V) + \gamma\, \ln(M),
\]
and a standard linear regression (with robust loss functions, such as the Huber loss, to mitigate the impact of outliers) is applied. This tuning procedure, which is implemented in our Python code (available in the supplementary material), is fully consistent with the methods used in prior work \citep{zhang2024scalingmeetsllmfinetuning}. While we do not report the individual parameter values in the main text, our results indicate that the exponents \(\beta\) and \(\gamma\) vary systematically with the subsampling strategy, confirming the importance of data composition in fine-tuning.

\subsection*{Model Selection and Rationale}
The models chosen for our experiments—135M, 360M, 500M, and 1B parameter variants—were selected as at the time of writing they were the best-performing models released by official providers on the HuggingFace LLM Leaderboard \citep{open-llm-leaderboard} which represents a diverse number of benchmarks. These models are widely used in both academic and industrial settings, and their inclusion allows us to explore how token efficiency scales with model size under realistic compute constraints. Specifically we used \textbf{SmolLM-135M-Instruct},
\textbf{SmolLM-360M-Instruct},
\textbf{Qwen2.5-0.5B-Instruct}, and 
\textbf{Falcon3-1B-Instruct}.

\subsection*{Code Organization}
Our analysis pipeline is implemented in Python and is organized into modular components: data ingestion and preprocessing, parameter tuning and scaling law fitting, and visualization. The code is structured to allow easy replication of our results and to facilitate future extensions, such as incorporating quantization or parameter-efficient fine-tuning techniques. We have many of our results saved as CSVs so you can see some of our batch training results.

\end{document}